\documentclass[journal]{IEEEtran}

\usepackage{cite}
\usepackage[colorlinks=true, linkcolor=black, citecolor=black, urlcolor=black]{hyperref}
\usepackage{amsmath}
\usepackage{amssymb}
\usepackage{booktabs}
\usepackage{multirow}
\usepackage{afterpage}
\usepackage{xcolor}
\usepackage[caption=false,font=footnotesize]{subfig}
\usepackage{dblfloatfix}

\ifCLASSINFOpdf
  \usepackage[pdftex]{graphicx}

\else

\fi

\begin{document}

\title{Uni-RCM: Unified Reference-guided Cross-modal Mapping for Multi-Class Anomaly Detection}

\author{Yangchen~Wu and 
        Huiqiang~Xie,~\IEEEmembership{Member,~IEEE}
\thanks{Y. Wu and H. Xie are with the School of Information Science and Technology, Jinan University, Guangzhou, Guangdong 510632, P. R. China (e-mail: wuyangchen@stu2023.jnu.edu.cn, huiqiangxie@jnu.edu.cn).}

}

\maketitle

\begin{abstract}
Multi-modal industrial anomaly detection typically relies on separate models for each product category, fundamentally limiting practical scalability. When shifting to a unified paradigm that handles diverse classes simultaneously, detection accuracy often degrades due to inter-class interference and feature manifold confusion. To overcome these challenges, we propose a Unified Reference guided Cross-modal Mapping framework, named Uni-RCM. At its core, we propose a reference guide block to dynamically filter out category-specific noise by introducing a learnable reference feature, which captures the commonalities across different modalities. Besides, an offline residual quantizer is proposed to characterize the normal distribution by multiple cascaded codebooks. Extensive evaluations on the MVTec-3D AD dataset demonstrate the state-of-the-art performance in the challenging multi-class setting and in terms of image-level detection and pixel-level localization.
\end{abstract}


\IEEEpeerreviewmaketitle

\section{Introduction}
\IEEEPARstart{I}{ndustrial} anomaly detection (IAD) is a critical technology for ensuring product quality in smart manufacturing. Powered by artificial intelligence, IAD has been shifting the paradigm from manual inspection to automated defect identification \cite{10144292,10262012}. In real-world scenarios, production lines need to handle diverse product categories. While conventional category-aware approaches rely on training individual models for specific categories, they inherently suffer from poor scalability and limited generalization to unseen classes. To overcome these limitations, category-agnostic models have gained significant attention, which aims to establish a shared feature space that captures the commonalities and variations across different product types.

Existing works for unified anomaly detection primarily focus on single-modality images~\cite{reference,uniad,omnial,uniflow,diffusion_anomaly}. Early efforts in this direction, You \textit{et al.}~\cite{uniad} have demonstrated the feasibility of detecting anomalies across multiple categories simultaneously within a single network. Building upon this paradigm, subsequent research has explored various architectures and modeling strategies to enhance feature representation. For instance, Zhao~\cite{omnial} has introduced OmniAL, which leverages a unified CNN framework for unsupervised anomaly localization. More recently, advanced generative techniques have been adapted to address the multi-class setting. Zhong and Song~\cite{uniflow} have proposed UniFlow, utilizing normalizing flows to estimate unified density distributions, while He \textit{et al.}~\cite{diffusion_anomaly} have developed a diffusion-based framework to reconstruct normal patterns across diverse classes. However, these unified models rely solely on 2D images, which fundamentally limits their capability to detect 3D structural deformations.

To overcome the inherent limitations of single-modality detection, multi-modal IAD methods have been proposed and are mainly categorized into memory-based \cite{mvtec3d,m3dm,g2sf,2m3df,HeZSY25,24spl} and reconstruction-based~\cite{23spl,ast,lsfa,cfm} methods. Following the foundational 3D anomaly detection baselines established by Bergmann \textit{et al.}~\cite{mvtec3d}, subsequent research has extensively explored multimodal feature fusion. For instance, Wang \textit{et al.}~\cite{m3dm} proposed a hybrid fusion framework to effectively combine 2D and 3D representations. Recent advancements in this vein further enhance detection robustness, such as the geometry-guided score fusion introduced by Tao \textit{et al.}~\cite{g2sf} and the multi-perspective fusion network developed by Asad \textit{et al.}~\cite{2m3df}. Conversely, reconstruction-based methods identify defects by measuring discrepancies between the original inputs and their reconstructed normal patterns or by distilling errors. Within this category, Rudolph \textit{et al.}~\cite{ast} utilized asymmetric student-teacher networks for anomaly distillation. Similarly, Costanzino \textit{et al.}~\cite{cfm} introduced cross-modal feature mapping to align distinct modalities, while Tu \textit{et al.}~\cite{lsfa} proposed self-supervised feature adaptation to optimize 3D industrial anomaly detection. However, these works focus on the category-aware design and omit the multi-class scenarios.

\begin{figure*}[!t]
  \centering
  \includegraphics[width=\textwidth]{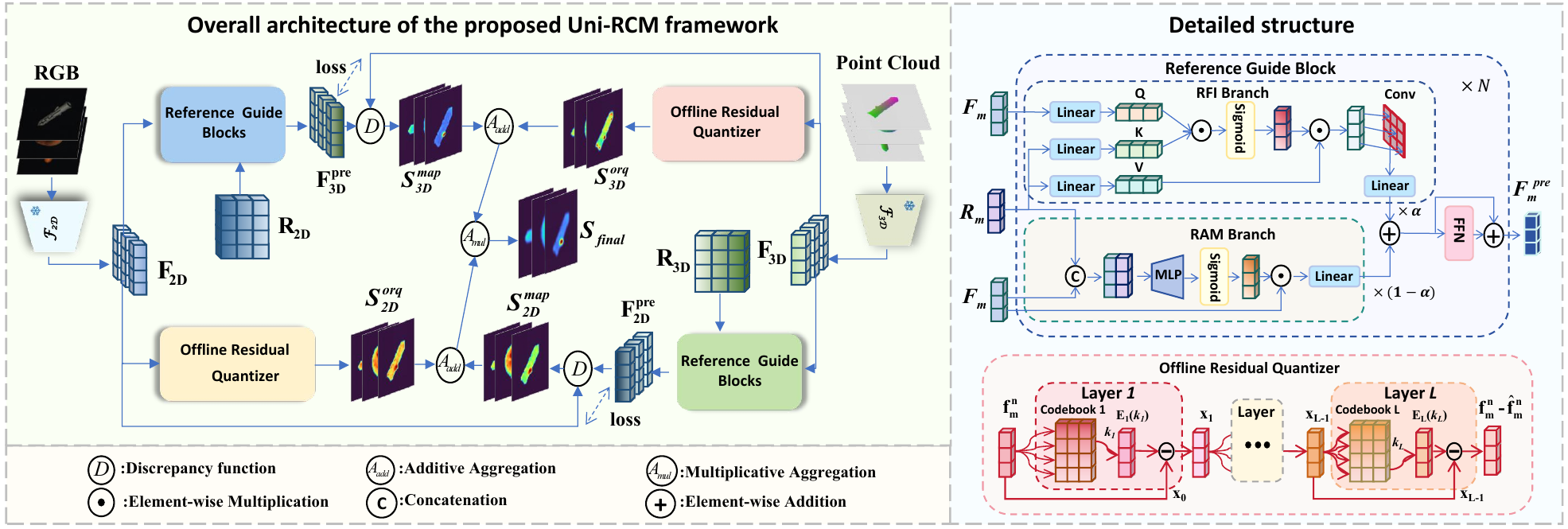}
  \caption{Overview of the proposed Uni-RCM framework, with the detailed structure of its core modules illustrated.}
  \label{fig:architecture_main}
\end{figure*}

When directly extending these methods to multi-class scenarios, multimodal anomaly detection accuracy faces degradation. The performance degradation stems from two fundamental limitations. First, reconstruction-based models are tailored to fit the normal distribution of individual categories, and multi-class data disrupts this distribution consistency.  Besides, the multi-class features cause memory banks to suffer from severe manifold confusion during retrieval, which leads to inevitable false positives.

To address the aforementioned limitations, we propose Uni-RCM, a Unified Reference-guided Cross-modal Mapping framework for multi-modal multi-class IAD. To mitigate category discrepancies, we design a reference guide module, which includes reference feature integration (RFI) and reference-guided attention modulation (RAM) to dynamically extract modality-consistent features. Furthermore, to circumvent manifold confusion during memory retrieval, we propose an offline residual quantizer. This module explicitly models compact normal pattern distributions, providing highly discriminative and complementary anomaly scores. The main contributions of this letter are summarized as follows: 
\begin{itemize}
    \item We propose the Uni-RCM framework for the multi-modal multi-class IAD task, which can effectively mitigate 2D and 3D feature confusion across multiple categories.  
    \item We design a reference guide module to dynamically regularize cross-modal mapping, and a offline residual quantizer module to provide robust auxiliary anomaly scoring.
    \item Extensive experiments on the MVTec-3D AD dataset demonstrate that Uni-RCM achieves state-of-the-art I-AUROC and AUPRO@1\% scores.
\end{itemize}

\section{Method}
\subsection{Network Overall Architecture}
As illustrated in Fig.~\ref{fig:architecture_main}, the proposed Uni-RCM framework employs two frozen pre-trained models to extract heterogeneous features $\mathbf{F}_m = [\mathbf{f}^{1}_m,\mathbf{f}^{2}_m, \cdots, \mathbf{f}^{N}_m], m\in\{2 D, 3D\}$, including the DINO ViT-B/8~\cite{dino} trained on ImageNet~\cite{imagenet} for RGB images and the PointMAE~\cite{pointmae} trained on ShapeNet~\cite{shapenet} for point clouds. The network then utilizes two key modules. One is the reference guided block for cross-modal feature alignment, which comprises RFI and RAM branches. Another is an offline residual quantizer, which is designed to store normal patterns and provide auxiliary anomaly signals. Next, we will introduce the two modules in detail.

\subsection{Reference Guide Block}
As detailed in Fig.~\ref{fig:architecture_main}, the reference guide block introduces learnable reference, $\mathbf{R}_m,m\in\{2D, 3D\}$, as the shared features. Taking input features $\mathbf{F}_m$ and $\mathbf{R}_m$ as inputs, the block performs feature interaction through two parallel branches, i.e., RFI and RAM branches. The RFI branch uses $\mathbf{F}_m$ as a query to retrieve relevant information from $\mathbf{R}_m$ by
\begin{equation}
\mathbf{Z}_{\text{RFI}} = f_\text{Conv}\left( \sigma\left(\frac{\mathbf{Q} \odot \mathbf{K}}{\sqrt{C}}\right) \odot \mathbf{V} \right),
\end{equation}
where $\mathbf{Z}_{\text{RFI}}$ is the retrived features, $f_\text{Conv}(\cdot)$ is the convolution network, $\sigma(\cdot)$ is the Sigmoid function, $\odot$ denotes element-wise multiplication, $C$ is constant, and $\mathbf{Q},\mathbf{K},\mathbf{V}$ are obtained via linear projections by $ \mathbf{Q} = \mathbf{W}_q \mathbf{F}_m$, $\mathbf{K} = \mathbf{W}_k \mathbf{R}_m$, and $\mathbf{V} = \mathbf{W}_v \mathbf{R}_m$.

Simultaneously, the RAM branch adaptively modulates $\mathbf{F}_m$ along the channel dimension. ${\mathbf F}_m$ and ${\mathbf R}_m$ are concatenated and fed into a multiple-layer perception (MLP) to generate modulated features by
\begin{equation}
\mathbf{Z}_{\text{RAM}} = \mathbf{F}_m \odot \sigma\left( f_{\text{MLP}}([\mathbf{F}_m; \mathbf{R}_m]) \right),
\end{equation}
where $\mathbf{Z}_{\text{RAM}}$ is the modulated features and $f_{\text{MLP}}(\cdot)$ is MLP function. 

To balance the importance between retrieved and the modulated features, a learnable scalar $\alpha$ is employed for dynamic weighted fusion. The fused feature is further processed by a feed-forward network with a residual connection to produce the final output $\mathbf{F}^{pre}$ by
\begin{equation}
\mathbf{F}^{pre}_m = f_\text{FFN}(\mathbf{Z}_{\text{fuse}}) + \mathbf{Z}_{\text{fuse}},
\end{equation}
where  $\mathbf{Z}_{\text{fuse}} = \alpha \cdot \mathbf{W_{\text{RFI}}Z}_{\text{RFI}} + (1 - \alpha) \cdot \mathbf{W_{\text{RAM}}Z}_{\text{RAM}}$ and $f_{\text{MLP}}(\cdot)$ is feed-forward network.

\textbf{\textit{Remark:}}  The insights behind the design of the reference-guided block can be summarized as follows. First, the learnable reference is used to store the common normal characteristics within cross-modal mapping. Then, the attention mechanism in the RFI branch helps the inputs to retrieve normal cross-modal details from the learnable reference, which makes the outputs closer to the normal cross-modal features. Besides, the RAM branch fuses the reference information to adaptively filter out abnormal details and enhance the normal details, effectively mitigating feature interference caused by multi-class mixtures. By smoothly coordinating these two mechanisms via dynamic weights, the single network achieves stable and high-quality cross-modal mapping that outputs the normal features with abnormal inputs.

\subsection{Offline Residual Quantizer}
To build a compact space of normal features, we introduce the offline residual quantizer (ORQ). Unlike memory bank approaches that require massive storage, ORQ offers a highly lightweight alternative through residual vector quantization. 

We define codebooks as a finite set $\mathbf {\mathcal{C}} = {\{\mathcal{C}_1, \mathcal{C}_2, \cdots, \mathcal{C}_L}\}$, in which $\mathcal{C}_l = \{(k_l, {\bf E}_l(k_l))\}_{k_l \in [K_l]}$, where each discrete index $k_l$ is mapped to an vector ${\bf E}_l(k_l)$. Here, $K_l$ represents the total capacity of the $\mathcal{C}_l$. 

Given the $\mathbf{f}^{n}_m$, ORQ employs $L$ cascaded codebooks to approximate it from coarse to fine. Starting with $0$-th residual ${\bf x}_0 = {\bf f}^n_m$, the $l$-th residual ${\bf x}_l$ is given by
\begin{equation}\label{eq-4}
\begin{aligned}
    k_l^i &= \arg \min\limits_{k_l \in [K_l]} \|\mathbf{x}_{l-1} - \mathbf{E}_l(k_l)\|^2, \\
    {\bf x}_l &= {\bf x}_{l-1} - {\bf E}_l(k_l^i),
\end{aligned}
\end{equation}
where $k_l^i $ is the index of the closest feature in ${\bf E}_l$ to $\mathbf{x}_{l-1}$. Therefore, the final reconstructed feature is $\hat{\mathbf{f}}^n_m = \sum_{l=1}^L {\bf E}_l(k_l^i )$. Then, by approximating all features in $\mathbf{F}_m$ by ORQ, we can obtain the $\hat{\mathbf{F}}_m$.

This achieves high-precision coverage of the normal feature space with a tiny codebook size. To prevent codebook collapse, the codebooks are generated by standard K-Means clustering on training features offline.

\textbf{\textit{Remark:}} The insights behind this module are summarized as follows. Due to tiny surface defects that belong to the high-frequency information, quantization is an effective way to perform a low-pass filter and filter out the defects, such that the outputs are smooth features without defects and are used to find deviations from the expected manifold. Besides, compared to conventional unified memory banks that need an additional memory space for each anomaly category, the residual quantization can approximate any anomaly types by the cascaded codebook design. Finally, this cascaded quantization effectively decomposes coarse category semantics and fine intra-class details, which is highly suited for the multi-class scenarios.

\subsection{Loss Functions and Joint Training}
To optimize Uni-RCM, we employ a joint loss function combining cosine similarity ($\mathcal{L}_{\text{cos}}$) and mean squared error ($\mathcal{L}_{\text{mse}}$) to align the generated features with the target modality in both direction and magnitude by
\begin{align}
\mathcal{L}_{\text{cos}} &= 1 -  \sum_{m \in \{2D, 3D\}} \frac{{\mathbf F}_{m} \cdot {\mathbf F}_{m}^{pre}}{\|{\mathbf F}_{m}^{}\|_2 \|{\mathbf F}_{m}^{pre}\|_2}, \\
\mathcal{L}_{\text{mse}} &=  \sum_{m \in \{2D, 3D\}} \|{\mathbf F}_{m}^{} - {\mathbf F}_{m}^{pre}\|_2^2.
\end{align}
The final training objective is the weighted sum $\mathcal{L}_{\text{total}} = \mathcal{L}_{\text{cos}} + \lambda \mathcal{L}_{\text{mse}}$, where $\lambda$ is a balancing coefficient (set to 0.01). The network adaptively balances the mapping difficulty between different modalities via the learnable parameter $\alpha$ in the reference-guided blocks, without requiring additional constraints.

\subsection{Anomaly Inference}

During inference, Uni-RCM detects anomalies through bidirectional mapping consistency and ORQ reconstruction errors. The mapping anomaly score $\mathcal{S}_{\text{map}}^{m}$ is measured by the Euclidean distance between the mapped and target features:
\begin{equation}
\mathcal{S}^{\text{map}}_{m} = \|{\mathbf F}_{m} - {\mathbf F}_{m}^{pre}\|_2, \quad m \in \{2D, 3D\}.
\end{equation}
Simultaneously, the quantization anomaly score uses the ORQ residual magnitude $\mathcal{S}^{\text{orq}}_{m} = \|\mathbf{F}_{m} - \hat{\mathbf{F}}_{m}\|_2$. To fully exploit multi-modal complementary information, the final pixel-level anomaly score $\mathcal{S}_{\text{final}}$ is defined as the multiplicative fusion of these signals:
\begin{equation}
\mathcal{S}_{\text{final}} = (\mathcal{S}^{\text{map}}_{2D} + \beta \mathcal{S}^{\text{orq}}_{2D}) \odot (\mathcal{S}^{\text{map}}_{3D} + \beta \mathcal{S}^{\text{orq}}_{3D}),
\end{equation}
where $\beta$ is a balancing coefficient (set to 0.1).
\section{Experiments}
\subsection{Datasets and Evaluation Metrics}
We evaluate Uni-RCM on the MVTec-3D AD dataset~\cite{mvtec3d}, which consists of 10 industrial product categories with various surface and structural defects. The model is trained on normal samples and tested on both normal and anomalous instances. Performance is assessed using Image-level AUROC (I-AUROC) for detection and Area Under the Per-Region Overlap curve (AUPRO@30\% and AUPRO@1\%) for segmentation.

\begin{table*}[!t]
  \centering
  \renewcommand{\arraystretch}{1.3}
  \caption{Anomaly detection performance comparison (I-AUROC and AUPRO@30\%). The \textbf{best} and \underline{second best} results in the multi-class setting are highlighted.}
  \label{tab:performance_comparison}
  \resizebox{\textwidth}{!}{
  \setlength{\tabcolsep}{2.5pt}
  \begin{tabular}{cl|ccccccccccc}
  \hline
  \hline
  \multicolumn{2}{c|}{\textbf{Method}} & Bagel & C\_Gland & Carrot & Cookie & Dowel & Foam & Peach & Potato & Rope & Tire & \textbf{Mean} \\ \hline
			& BTF~\cite{btf} & 0.819\,\color{gray}{/0.975} & 0.742\,\color{gray}{/0.957} & 0.965\,\color{gray}{/0.979} & 0.864\,\color{gray}\underline{/0.972} & 0.956\,\color{gray}\underline{/0.969} & 0.568\,\color{gray}{/0.863} & 0.825\,\color{gray}{/0.974} & 0.884\,\color{gray}{/0.981} & 0.933\,\color{gray}{/0.927} & 0.852\,\color{gray}{/0.935} & 0.841\,\color{gray}{/0.953} \\
			& AST~\cite{ast} & 0.754\,\color{gray}{/0.946} & 0.599\,\color{gray}{/0.795} & 0.469\,\color{gray}{/0.942} & 0.761\,\color{gray}{/0.813} & 0.624\,\color{gray}{/0.844} & 0.628\,\color{gray}{/0.567} & 0.470\,\color{gray}{/0.925} & 0.499\,\color{gray}{\underline{/0.982}} & 0.590\,\color{gray}{/0.707} & 0.682\,\color{gray}{/0.695} & 0.608\,\color{gray}{/0.822} \\
			& M3DM~\cite{m3dm} & 0.835\,\color{gray}{/0.893} & 0.644\,\color{gray}{/0.896} & 0.795\,\color{gray}{/0.957} & 0.850\,\color{gray}{/0.840} & 0.708\,\color{gray}{/0.910} & 0.759\,\color{gray}{/0.907} & 0.760\,\color{gray}{/0.926} & 0.837\,\color{gray}{/0.925} & 0.843\,\color{gray}{/0.908} & 0.651\,\color{gray}{/0.837} & 0.768\,\color{gray}{/0.900} \\
			& CFM~\cite{cfm} & 0.978\,\color{gray}{/0.975} & \underline{0.880}\,\color{gray}\underline{/0.967} & 0.940\,\color{gray}{\textbf{/0.982}} & \textbf{1.000}\,\color{gray}{/0.953} & \underline{0.998}\,\color{gray}{/0.962} & 0.878\,\color{gray}{/0.958} & 0.919\,\color{gray}{/0.979} & 0.888\,\color{gray}\underline{/0.982} & 0.977\,\color{gray}\textbf{/0.979} & \underline{0.860}\,\color{gray}{\underline{/0.978}} & \underline{0.932}\,\color{gray}\underline{/0.972} \\
			& LSFA~\cite{lsfa} & 0.942\,\color{gray}{/0.974} & 0.791\,\color{gray}{/0.962} & 0.919\,\color{gray}{/0.980} & 0.952\,\color{gray}{{/0.958}} & 0.932\,\color{gray}\textbf{/0.984} & 0.818\,\color{gray}{{/0.963}} & \underline{0.966}\,\color{gray}\textbf{/0.989} & 0.657\,\color{gray}{/0.967} & \textbf{0.985}\,\color{gray}{/0.972} & 0.783\,\color{gray}{/0.954} & 0.875\,\color{gray}{/0.970} \\
			& G$^2$SF~\cite{g2sf} & \textbf{0.993}\,\color{gray}{\textbf{/0.982}} & 0.855\,\color{gray}{\textbf{/0.974}} & \textbf{0.998}\,\color{gray}{\textbf{/0.982}} & 0.907\,\color{gray}\textbf{/0.973} & 0.915\,\color{gray}\underline{/0.969} & \textbf{0.944}\,\color{gray}{\textbf{/0.973}} & \textbf{0.978}\,\color{gray}{\underline{/0.982}} & \textbf{0.969}\,\color{gray}{\underline{/0.982}} & 0.967\,\color{gray}{/0.976} & 0.749\,\color{gray}{/0.971} & 0.928\,\color{gray}{\textbf{/0.976}} \\
			& CIF~\cite{cif} & {0.849}\,\color{gray}{/0.596} & {0.717}\,\color{gray}{/0.808} & {0.749}\,\color{gray}{/0.936} & {0.771}\,\color{gray}{/0.570} & {0.672}\,\color{gray}{/0.906} & {0.816}\,\color{gray}{/0.679} & {0.726}\,\color{gray}{/0.713} & {0.512}\,\color{gray}{/0.899} & {0.917}\,\color{gray}{/0.834} & {0.602}\,\color{gray}{/0.833} & {0.733}\,\color{gray}{/0.777} \\
			& \textbf{Ours} & \underline{0.988}\,\color{gray}\underline{/0.977} & \textbf{0.897}\,\color{gray}{/0.966} & \underline{0.966}\,\color{gray}\textbf{/0.982} & \underline{0.999}\,\color{gray}{{/0.948}} & \textbf{1.000}\,\color{gray}{{/0.954}} & \underline{0.898}\,\color{gray}\underline{/0.965} & 0.963\,\color{gray}{{/0.981}} & \underline{0.950}\,\color{gray}\textbf{/0.983} & \underline{0.979}\,\color{gray}\textbf{/0.979} & \textbf{0.903}\,\color{gray}\textbf{/0.982} & \textbf{0.954}\,\color{gray}\underline{{/0.972}} \\ 
  \hline
  \end{tabular}
  }
  \end{table*}

\subsection{Implementation Details}
Data preprocessing follows \cite{m3dm}. Uni-RCM is implemented in PyTorch and trained on a single NVIDIA RTX 4090D GPU. We stack 3 reference-guided blocks with a hidden dimension of 512. The main network is trained end-to-end using the Adam optimizer for 200 epochs, with a batch size of 32 and a learning rate of $3 \times 10^{-4}$. In contrast, the ORQ module utilizes a 4-layer residual cascade with codebook sizes set to $K_{2D}=4096$ and $K_{3D}=1024$. During inference, the resultant anomaly maps are bilinearly interpolated to $224 \times 224$. Finally, a Gaussian smoothing filter following \cite{sub} is applied to smooth noise.

\begin{table}[ht]
\renewcommand{\arraystretch}{1.1}
\caption{AUPRO@1\% Results. The \textbf{best} and \underline{second best} results in the multi-class setting are highlighted.}
\label{tab:aupro_1}
\centering
\resizebox{\columnwidth}{!}{
\setlength{\tabcolsep}{2.5pt}
\begin{tabular}{lccccccc}
\toprule
\textbf{Metric} & BTF~\cite{btf}& AST~\cite{ast}& M3DM~\cite{m3dm}& CFM~\cite{cfm} &LSFA~\cite{lsfa}& G$^2$SF~\cite{g2sf}& \textbf{Ours} \\ \midrule
AUPRO@1\% & 0.366 & 0.303 & 0.192 & 0.442  &0.408& \underline{0.448} & \textbf{0.455} \\ \bottomrule
\end{tabular}
}
\end{table}

\begin{figure*}[!t]
\centering
\includegraphics[width=\textwidth]{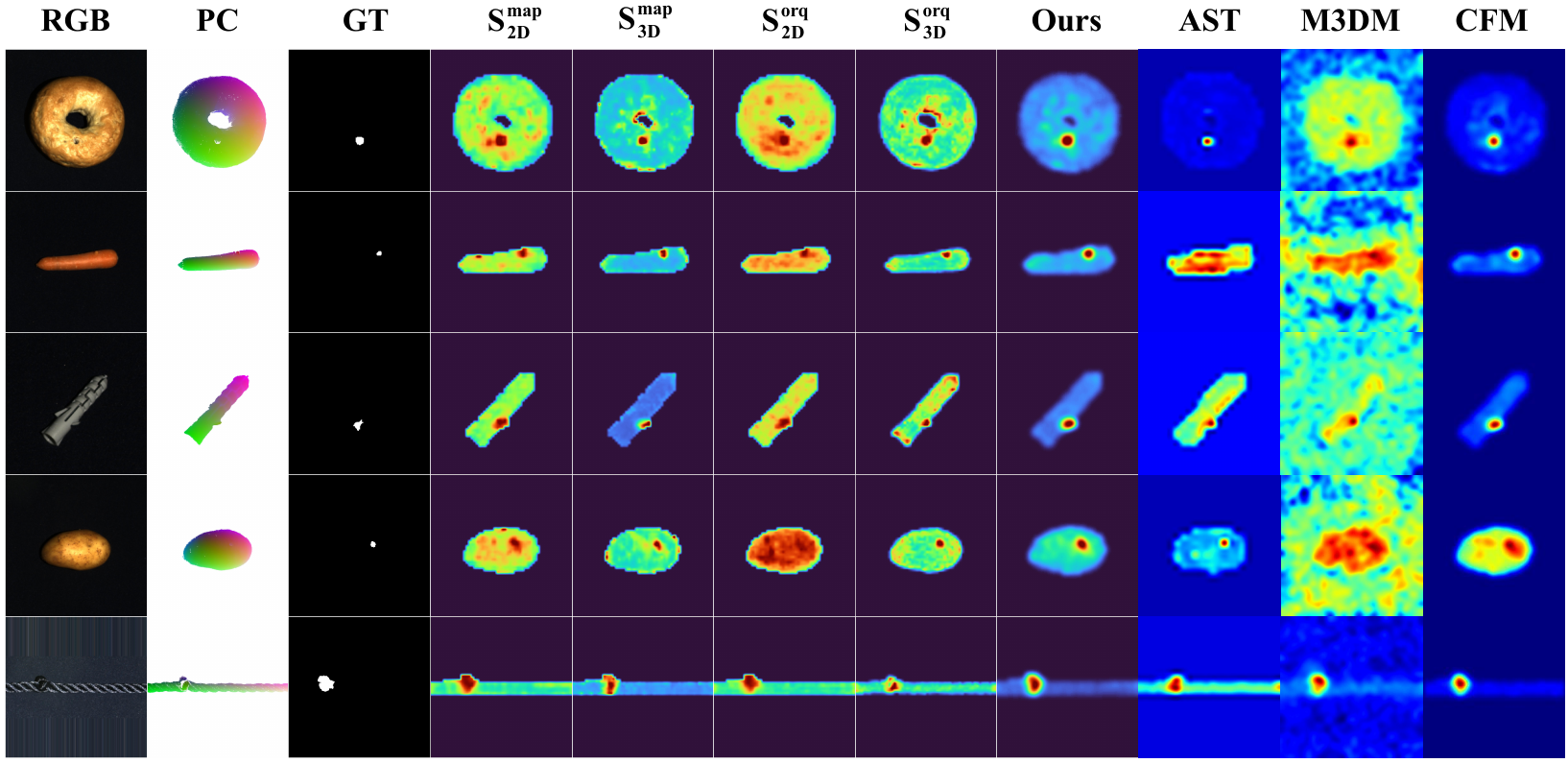}
\vspace{-0.5em}
\caption{Qualitative segmentation results. The left columns illustrate the internal anomaly-map generation process of Uni-RCM, including 2D/3D mapping anomaly scores and quantization anomaly scores. The right columns present a comparison of the final anomaly localization results with state-of-the-art methods.}\label{fig:qualitative_main}
\end{figure*}

\subsection{Comparison With Existing Methods}
Table~\ref{tab:performance_comparison} and Table~\ref{tab:aupro_1} compare Uni-RCM with state-of-the-art anomaly detection methods on the MVTec-3D AD dataset~\cite{mvtec3d} for a multi-class scenario. For the anomaly detection, Uni-RCM achieves state-of-the-art performance, outperforming previous methods by a margin of 2.2\% absolute points for I-AUROC. This indicates that the reference-guided mapping effectively manages inter-class interference for generalized detection. Regarding localization, Uni-RCM also outperforms all previous methods except for the $G^2$SF~\cite{g2sf} model that uses a memory bank. However, Uni-RCM achieves the best results on the stricter AUPRO@1\% metric with 0.7\% gain, which confirms that combining cross-modal mapping with ORQ effectively suppresses background noise and enhances boundary delineation. Qualitative results in Fig.~\ref{fig:qualitative_main} further corroborate this finding, where Uni-RCM produces sharper and more accurate anomaly maps compared to existing methods.

\subsection{Ablation Study}
To evaluate the contribution of proposed components in Uni-RCM, we conducted comprehensive ablation studies as detailed in Table~\ref{tab:ablation}. First, using either the RFI or RAM branch alone yields I-AUROC of 0.938 and 0.942, respectively. Fusing both improves it to 0.947, demonstrating that the learnable reference features obtain the general representations to manage multi-class variations. Besides, adding ORQ to the baseline mapping network improves I-AUROC from 0.947 to 0.954. This confirms that quantization reconstruction error provides a crucial complementary signal to constrain the normal feature manifold and improve anomaly separability.

\begin{table}[!t]
  \renewcommand{\arraystretch}{1.1}
  \caption{Ablation Study on Key Components in Uni-RCM. The \textbf{best} and \underline{second best} results in the multi-class setting are highlighted.}
  \label{tab:ablation}
  \centering
  \footnotesize
  \setlength{\tabcolsep}{4pt}
  \begin{tabular}{ccc|ccc}
  \hline
  \hline
  ORQ& RFI& RAM& I-AUROC & AUPRO@30\%  &AUPRO@1\%\\ \hline
  \checkmark & & & 0.783& 0.954&    0.406\\
  & \checkmark & & 0.938& 0.969&    0.449\\
 & & \checkmark& 0.942& 0.970&0.450\\
  & \checkmark & \checkmark& \underline{0.947} & \textbf{0.972}  &\underline{0.452}\\
  \checkmark & \checkmark & \checkmark & \textbf{0.954} & \textbf{0.972}  &\textbf{0.455}\\ \hline
  \hline
  \end{tabular}
  \end{table}

\section{Conclusion}
In this letter, we proposed Uni-RCM, a Unified Reference-guided Cross-modal Mapping framework, to tackle multi-modal multi-class industrial anomaly detection task. To mitigate severe interference across diverse category distributions, we designed a reference guide block that dynamically aligns cross-modal features via learnable reference features and adaptive channel-wise modulation. Furthermore, we proposed an offline residual quantizer that decouples normal pattern memory by using compact cascaded codebooks. Extensive experiments on the MVTec-3D AD dataset demonstrate that Uni-RCM establishes a robust unified model, achieving state-of-the-art detection and fine-grained localization performance under the multi-class setting.

\ifCLASSOPTIONcaptionsoff
  \newpage
\fi

\bibliographystyle{IEEEtran}

\bibliography{IEEEabrv,references}

\end{document}